# Efficient training for large-scale optical neural network using an evolutionary strategy and attention pruning

Zhiwei Yang[a,b], Zeyang Fan[a,b], Yihang Lai[a,b], Qi Chen[a,b], Tian Zhang[a,b,*], Jian Dai[a,b], and Kun Xu[a,b]

[a] *State Key Laboratory of Information Photonics and Optical Communications, Beijing University of Posts and Telecommunications, Beijing, 100876, China*
[b] *School of Electronic Engineering, Beijing University of Posts and Telecommunications, Beijing, 100876, China*
[*] *Corresponding author (ztian@bupt.edu.cn)*

*Abstract*—MZI-based block optical neural networks (BONNs), which can achieve large-scale network models, have increasingly drawn attentions. However, the robustness of the current training algorithm is not high enough. Moreover, large-scale BONNs usually contain numerous trainable parameters, resulting in expensive computation and power consumption. In this article, by pruning matrix blocks and directly optimizing the individuals in population, we propose an on-chip covariance matrix adaptation evolution strategy and attention-based pruning (CAP) algorithm for large-scale BONNs. The calculated results demonstrate that the CAP algorithm can prune 60% and 80% of the parameters for MNIST and Fashion-MNIST datasets, respectively, while only degrades the performance by 3.289% and 4.693%. Considering the influence of dynamic noise in phase shifters, our proposed CAP algorithm (performance degradation of 22.327% for MNIST dataset and 24.019% for Fashion-MNIST dataset utilizing a poor fabricated chip and electrical control with a standard deviation of 0.5) exhibits strongest robustness compared with both our previously reported block adjoint training algorithm (43.963% and 41.074%) and the covariance matrix adaptation evolution strategy (25.757% and 32.871%), respectively. Moreover, when 60% of the parameters are pruned, the CAP algorithm realizes 88.5% accuracy in experiment for the simplified MNIST dataset, which is similar to the simulation result without noise (92.1%). Additionally, we simulationally and experimentally demonstrate that using MZIs with only internal phase shifters to construct BONNs is an efficient way to reduce both the system area and the required trainable parameters. Notably, our proposed CAP algorithm show excellent potential for larger-scale network models and more complex tasks.

## 1. Introduction

Owing to the high computational speed, large bandwidth, low energy consumption and low latency, the emerging optical neural networks (ONNs) have the potential to break through the bottlenecks about speed and energy consumption of electric computing hardware, attracting extensive attentions [1]. MZI-based ONNs [2], one of the mainstream architectures of ONNs, are particularly valued for their high integration density, configurability and high stability, making them extensively applied. Many earlier MZI-based ONNs were typically configured to handle tasks such as matrix multiplication or convolution [3-5], but these networks were often constrained to small-scale models. This is because that the reported single integrated photonic chip can only achieve about 128 ports at the current state-of-art [6]. However, such chip still can't meet the requirements of complex tasks. For example, without data preprocessing, simple tasks like MNIST typically require the number of ports to reach around one thousand [7], and complex tasks like CIFAR-10 require thousands or even tens of thousands of ports [8]. Therefore, studying the method of using small-scale optical chip to complete complex tasks is significant.

One practical approach to achieving large-scale ONNs is dividing the weight matrix of neural networks into multiple small blocks [9, 10], thereby reducing the required scale of the photonic chip. This technique has already been successfully implemented in electronic artificial neural networks (ANNs) [11, 12]. However, the training algorithms for these large-scale block optical neural networks (BONNs) are still not sufficiently effective. On one hand, although the finite-difference-based gradient training approach [13] and the stochastic gradient descent (SGD)-based training algorithm [2], which have been reported for small-scale ONNs, can be directly extended to large-scale BONNs, they have the disadvantage of easily converging to local optima due to excessive reliance on initial parameter settings [13] and mapping errors between pre-trained weights and the actual phase values caused by noise in photonic chips [2], respectively. On the other hand, several gradient-based on-chip training methods for large-scale BONNs have been reported, including the three-stage learning framework called L²ight [9] and the block adjoint training (BAT) algorithm [14]. However, gradient information can only reflect local changes other than global changes, making gradient-based algorithms difficult to converge to the global optimum. Moreover, the noise in photonic chips can interfere with gradient calculation and make the gradient direction inaccurate, ultimately leading to the gradient-based algorithms with a weak robustness. Furthermore, these issues will become exacerbated in large-scale BONNs.

In contrast to gradient-based training algorithms, by maintaining a diverse population, evolutionary strategies, which don't rely on gradient information of the objective function, can effectively avoids getting stuck in local optima and are suitable to optimize problems with noise [15]. Therefore, evolutionary strategies have the potential to exhibit strong robustness in large-scale BONNs which are affected by noise. Our team has previously employed genetic algorithm and particle swarm optimization [16] to design and train ONNs. However, these methods, which involve updating dozens to hundreds of individuals, tend to be limited by the large time-consuming, especially for large-scale BONNs. As a standout method in evolutionary strategies, the covariance matrix adaptation evolution strategy (CMA-ES) [17] can capture the correlation between variables by introducing an adaptive adjustment covariance matrix to reduce the dimensionality of the search space and to guide the search direction, and can also adaptively adjust the search step size. Therefore, the CMA-ES has excellent ability to solve high-dimensional optimization problems and only requires a relatively small population to quickly converge to the global optimum, making it a highly attractive option for training ANNs [18] and designing photonic devices [19]. Notably, the scale expansion of BONNs will increase the number of trainable parameters, significantly escalating the computational costs of CMA-ES [20]. Therefore, researching effective pruning techniques for CMA-ES and BONN is significant.

Therefore, in this paper, by pruning blocks that have minimal impact on the results and directly optimizing the individuals in the population, we propose a novel on-chip CMA-ES and attention-based pruning (CAP) algorithm for large-scale BONNs. As the research foundation of CAP algorithm, we analyze the effect of the block size and population size on the performance of the CMA-ES based on BONNs. Moreover, we compare the performance of the BONNs with and without external phase shifters. For MNIST and Fashion-MNIST datasets, we evaluate the CAP algorithm against the random pruning-based CMA-ES (RPC) algorithm to highlight its effectiveness, and compare the CAP algorithm with our previously proposed BAT algorithm [14] and CMA-ES to demonstrate its robustness. In addition, we deploy the CAP algorithm on actual photonic chips and experimentally demonstrate its practicality. Notably, our proposed CAP algorithm show significant promise for larger-scale network models and more complex tasks.

## 2. Training method based on CAP algorithm

The algorithmic details of our proposed CAP algorithm, as shown in Fig. 1(a), are outlined as follows:

1. Initialize network architecture and parameters. Firstly, we establish a BONN with block size $g$ to complete block matrix multiplication for the target dataset. As shown in Fig. 1(b), the MZI-based BONN is composed of multiple block unitary matrix layers (BUML$_1$, ..., BUML$_l$, ..., BUML$_L$), and each BUML is constructed using a photonic integrated chip (PIC) with I/O and an array of photodetector (PD). For each BUML, loading the inputted signals, controlling the weight matrices, and monitoring the outputted signals from PD are implemented by electronic computer (EC). Typically, the output of each layer in the traditional ONNs can be mathematically described as $\mathbf{X}_l=f_l(\mathbf{W}_l\mathbf{X}_{l-1})$, where $\mathbf{X}_{l-1}$ ($l$=1, 2, ..., $L$), $\mathbf{X}_l$, $\mathbf{W}_l$ and $f_l$ are the inputted signal, outputted signal, weight matrix and nonlinear activation function of $l$-th layer of ONNs, respectively. Unlike inference process of the traditional ONNs, the weight matrix in BONN is divided into many smaller blocks. The computation results of each block undergo nonlinear transformations and are then processed following the principles of block matrix multiplication. This process can be described as:

$$X_{l,i} = f_l(W_{l,i1}X_{l-1,1}) + \cdots + f_l(W_{l,ij}X_{l-1,j}) + \cdots + f_l(W_{l,iq}X_{l-1,q}) \quad (1)$$

where
$$W_l = \begin{bmatrix} W_{l,11} & \cdots & W_{l,1j} & \cdots & W_{l,1q} \\ \vdots & \ddots & \vdots & \ddots & \vdots \\ W_{l,i1} & \cdots & W_{l,ij} & \cdots & W_{l,iq} \\ \vdots & \ddots & \vdots & \ddots & \vdots \\ W_{l,p1} & \cdots & W_{l,pj} & \cdots & W_{l,pq} \end{bmatrix} \quad (2)$$

$$X_l = \begin{bmatrix} X_{l,1} \\ \vdots \\ X_{l,i} \\ \vdots \\ X_{l,p} \end{bmatrix} \quad X_{l-1} = \begin{bmatrix} X_{l-1,1} \\ \vdots \\ X_{l-1,j} \\ \vdots \\ X_{l-1,q} \end{bmatrix} \quad (3)$$

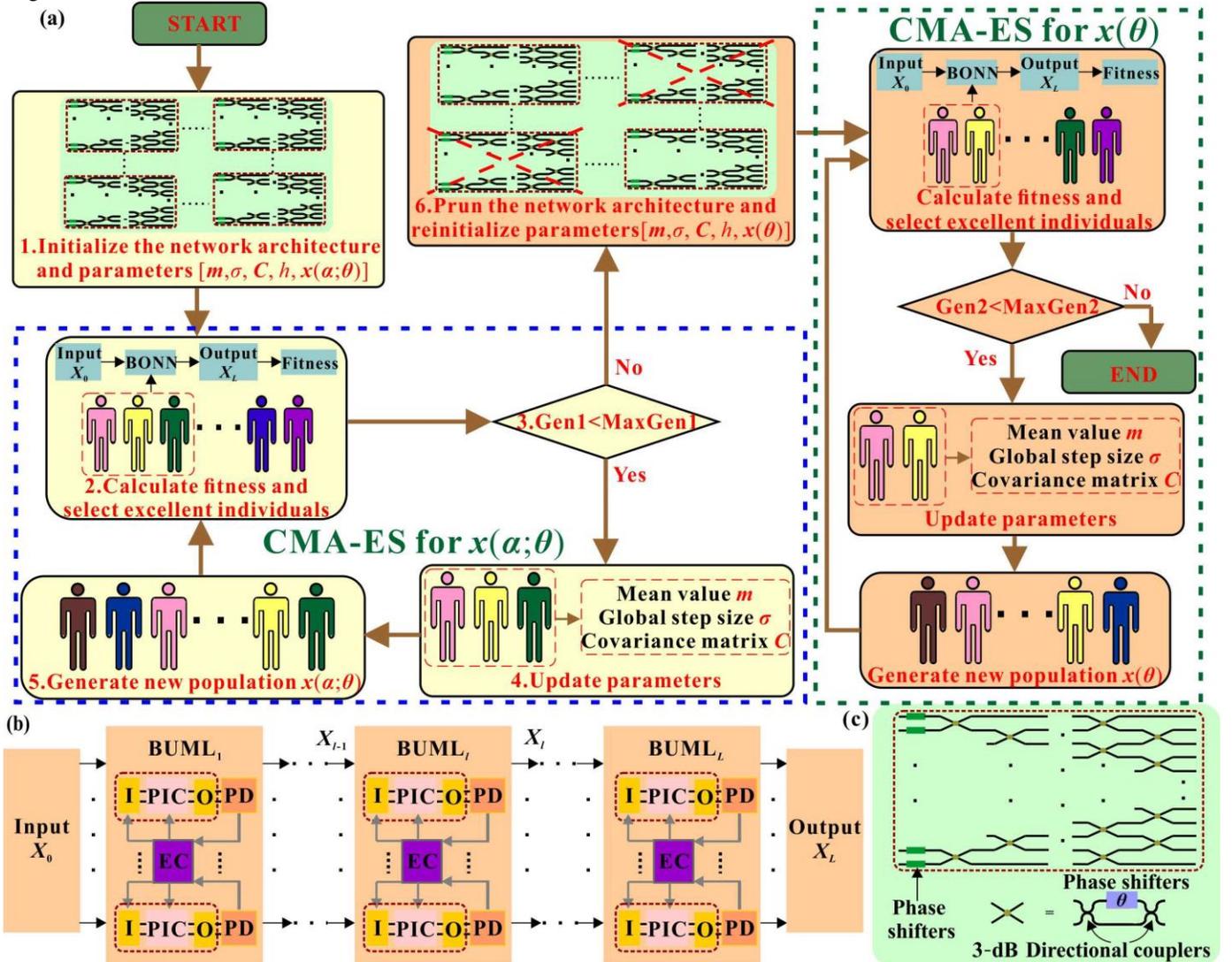

Fig. 1. (a) The flowchart of the training process for the BONN based on CAP algorithm. (b) The general scheme of the BONN based on MZIs. (c) The architecture of PIC.

Here, $\mathbf{W}_l$ is partitioned into $p \times q$ blocks of size $g \times g$, $\mathbf{W}_{l,ij}$ ($i$=1, 2, ..., $p$, $j$=1, 2, ..., $q$) is a block of weight matrix $\mathbf{W}_l$, $\mathbf{X}_{l,i}$ is the $i$-th row block of the outputted signal $\mathbf{X}_l$, $\mathbf{X}_{l-1,j}$ is the $j$-th row block of the inputted signal $\mathbf{X}_{l-1}$, respectively. In general, the arbitrary real-valued weight matrix can be decomposed into the product of two unitary matrices and one diagonal matrix according to the singular value decomposition principle [21]. However, using arbitrary real-valued weight matrix often results in higher computational complexity. Interestingly, neural networks without using complete weight matrices can achieve comparable performance to those using complete weight matrices [11]. Therefore, based on the decomposition method proposed by Clements et al. [22], we construct a PIC to achieve the block unitary matrix $\mathbf{W}_{l,ij}$, and reuse the PIC to complete each matrix-vector multiplication. In theory, our BONN with only block unitary matrix may experience performance degradation due to the reduced parameter space, but the degradation isn't significant [23].

As shown in Fig. 1(c), the PIC, simulated by Neuroptica [24], consists of an array of phase shifters and MZIs [25]. Typically, each MZI comprises an internal phase shifter, an external phase shifter and two 3-dB directional couplers. However, as the size of the BONN increases, the number of phase shifters required for the linear optical mesh increases significantly, leading to higher hardware costs. To mitigate this issue, we further remove the external phase shifters from the MZIs in the PIC which can only achieve block unitary matrix. The effectiveness of this method is demonstrated in the following part by comparing the performance of our BONN with that of the BONN with the external phase shifters. Moreover, photo-electronic conversion is required due to the post-processing of output signal of the PIC on the EC. Here, the PD array [26, 27] is used to complete photo-electronic conversion and can be regarded as a quadratic nonlinearity function. Importantly, the nonlinear function is applied to the outputted signal of each block, introducing more nonlinearity to BONN. This can enhance the ability of our proposed BONN to complete more complex tasks in some degrees.

Subsequently, the parameters, mainly including the population mean value $\mathbf{m}$, global step size $\sigma$, covariance matrix $\mathbf{C}$, population size $h$ and population individuals $\mathbf{x}$ are initialized. Specially, global step size $\sigma$>0 and covariance matrix $\mathbf{C}$ are initialized as 0.1 and an identity matrix, respectively. Then the $k$-th population individual of the first generation $\mathbf{x}_k^1$ is generated based on the initialized mean (search center), covariance matrix, and step size, and can be described by the following equation:

$$\begin{aligned} \mathbf{x}_k^{g+1} &\sim \mathbf{m}^g + \sigma^g N(0, \mathbf{C}^g) \\ k &= 1, 2, \cdots, h \end{aligned} \quad (4)$$

where $g$=0 and $N$ represents multivariate normal distribution. Here, a population individual $\mathbf{x}_k$ includes the attention coefficients $\boldsymbol{\alpha}$ and phase values of the phase shifters $\boldsymbol{\theta}$, and are usually represented by $\mathbf{x}_k(\boldsymbol{\alpha}; \boldsymbol{\theta})$. All the elements of $\boldsymbol{\alpha}$ vary from 0 to 1 and elements of $\boldsymbol{\theta}$ vary from 0 to $2\pi$. Notably, the attention coefficient $\boldsymbol{\alpha}$, which indicates the influence of the matrix block on the final output result, is assigned to each matrix block in BONN. Thus, Eq. (1) should be rewritten as:

$$X_{l,i} = a_{l,i1}f_l(W_{l,i1}X_{l-1,1}) + \cdots + a_{l,ij}f_l(W_{l,ij}X_{l-1,j}) + \cdots + a_{l,iq}f_l(W_{l,iq}X_{l-1,q}) \quad (5)$$

where
$$\boldsymbol{\alpha}_l = (a_{l,11}, \cdots, a_{l,ij}, \cdots, a_{l,pq}) \quad (6)$$

Here, $a_{l,ij}$ ($i$=1, 2, ..., $p$, $j$=1, 2, ..., $q$) is an element of attention coefficient $\boldsymbol{\alpha}_l$ of $l$-th layer of BONN and corresponds to the block weight matrix $\mathbf{W}_{l,ij}$.

2. Calculate fitness and select excellent individuals. Firstly, training datasets are preprocessed by using data enhancement method [28] to improve model generalization and to relieve overfitting. Subsequently, processed data is inputted into the BONN and transferred from the input layer to output layer. In the output layer, the predicted results for current iteration are collected, and the fitness value $f(\mathbf{x})$ between the prediction results $\mathbf{X}_L$ and target results $\mathbf{T}$ are calculated based on cross-entropy:

$$f(\mathbf{x}) = \sum_{i=1}^{p} \left( T_i \log \frac{e^{X_{L,i}(x)}}{\sum_{i=1}^{p} e^{X_{L,i}(x)}} \right) \quad (7)$$

where $X_{L,i}$ and $T_i$ represent elements of $\mathbf{X}_L$ and $\mathbf{T}$, respectively. Then the fitness values are sorted in descending order and the top $\mu = \lfloor h/2 \rfloor$ individuals with the highest fitness values are selected to participate in parameter updates. Here, $\lfloor \rfloor$ represents rounding downwards.

3. If the maximum generation is reached, execute step 6 and save the attention coefficients $\boldsymbol{\alpha}$. Otherwise, execute step 4.

4. Update parameters. The CMA-ES utilizes the selected population individuals to update population mean value $\boldsymbol{m}$, global step size $\sigma$, covariance matrix $\boldsymbol{C}$. Firstly, in order to move the population towards a better direction, $\mu$ excellent individuals are performed weighted-average operation to update $\boldsymbol{m}^{g+1}$:

$$\boldsymbol{m}^{g+1} = \sum_{i=1}^{\mu} \omega_i \boldsymbol{x}_i^{g+1} \quad (8)$$

where
$$\sum_{i=1}^{\mu} \omega_i = 1, \omega_1 \geq \omega_2 \cdots \geq \omega_\mu \geq 0 \quad (9)$$

Here $\omega_i$ represents the coefficient of influence on the next generation of individuals, $f(\boldsymbol{x}_1^{g+1}) \geq f(\boldsymbol{x}_2^{g+1}) \geq \cdots \geq f(\boldsymbol{x}_\mu^{g+1})$, respectively. Then covariance matrix $\boldsymbol{C}^{g+1}$ is updated to better adapt to the local characteristics of the objective function:

$$\boldsymbol{C}^{g+1} = (1 - c_1 - c_\mu) \boldsymbol{C}^g + c_1 \boldsymbol{p}_c^{g+1} (\boldsymbol{p}_c^{g+1})^T$$
$$+ c_\mu \sum_{i=1}^{\mu} \omega_i \frac{\boldsymbol{x}_i^{g+1} - \boldsymbol{m}^g}{\sigma^g} (\frac{\boldsymbol{x}_i^{g+1} - \boldsymbol{m}^g}{\sigma^g})^T \quad (10)$$

where $c_1$ and $c_\mu$ are constant learning rate, and are used to control the influence of evolutionary paths and population distribution, and $\boldsymbol{p}_c^{g+1}$ is used to record the direction of the mean's movement and is updated in every generation [17, 19]. Finally, global step size $\sigma^{g+1}$ is updated to control the search range of the population:

$$\sigma^{g+1} = \sigma^g e^{\frac{c_\sigma}{d_\sigma}\left(\frac{\|\boldsymbol{p}_\sigma^{g+1}\|}{E\|N(0,\boldsymbol{I})\|} - 1\right)} \quad (11)$$

where $d_\sigma \approx 1$, $c_\sigma = 2/(n+3)$, $n$ represents the number of variables in a population individual, $E\|N(0, \boldsymbol{I})\|$ represents the expectation of Euclidean norm of the $N(0, \boldsymbol{I})$ distribution vector, and $\boldsymbol{p}_\sigma^{g+1}$ is used to record changes in search direction and is updated in every generation [17, 19], respectively. Then the updated mean $\boldsymbol{m}^{g+1}$, global step size $\sigma^{g+1}$, and covariance matrix $\boldsymbol{C}^{g+1}$ are used to generate the population individuals for the next generation. More detail about the updating of parameters can be found in Ref. [17].

5. Generate new population. In each generation, new population is generated based on the current mean, covariance matrix, and step size. This process can be described by Eq. (4). Then execute step 2. Subsequently, repeat steps 4, 5, and 2 until reaching the maximum generation.

6. Prune network architecture and reinitialize parameters. We sort all the attention coefficients and prune the weight matrix blocks with smaller attention coefficients in the BONN initialized in step 1. Here, the number of pruned weight matrix blocks accounts for a predefined proportion $R$ of the total number of blocks. The pruned BONN is then utilized as a new network model to perform matrix multiplication following Eq. (1). Subsequently, the parameters, mainly including the population mean value $\boldsymbol{m}$, global step size $\sigma$, covariance matrix $\boldsymbol{C}$, population size $h$ and population individuals $\boldsymbol{x}$ are reinitialized. Here, population individuals only include the phase values of the phase shifters $\boldsymbol{\theta}$.

7. Update population individuals according to steps 2-5 until reaching the maximum generation.

## 3. Simulated and experimental results

### 3.1. Image Classification Based on CMA-ES

As an important component of the proposed CAP algorithm, we firstly only employ the CMA-ES for an unpruned BONN, as the green dashed line surrounded process shown in Fig. 1(a), and apply it to a standard MNIST dataset classification task [29] to decide the setting of the hyper-parameters, including the block size of BONN and population size $h$, in our proposed CAP algorithm. This dataset consists of 60000 training grayscale images with dimensions of 28 × 28 pixels (10 categories) and 10000 testing images. To classify these images, we construct a BONN with an architecture of 784-10 and use the CMA-ES to train this network. Notably, this BONN doesn't include the external phase shifters. Firstly, we study the impact of the different block size on the performance of the CMA-ES based on BONN as shown in Fig. 2(a). To consider the stability of the performance of the CMA-ES, avg10acc, which is the average test accuracy in the last 10 generations, is used. It can be found from Fig. 2(a) that the avg10acc of the BONN with 12 block size can reach 93.185%. This suggests that the proposed CMA-ES is effective for the BONN. Moreover, as seen in Fig. 2(a), we can also find that the avg10acc of the BONN with 4 block size, 8 block size, 10 block size and 22 block size can only reach 89.346%, 92.697%, 92.719%, and 92.848%, respectively. This suggests that different block size can influence the performance of the CMA-ES, and the BONN with 12 block size has the best performance. This characteristic can be attributed to the fact that number of PD utilized in the BONN diminishes as the block size increases, which leads to a decrease in the nonlinear capacity of the BONN. As is widely recognized, nonlinearity is crucial to the expressive power of neural networks. Weak nonlinearity can lead to the neural networks being able to only fit functions that are simpler than the objective distribution. On the other hand, excessive nonlinearity can reduce the generalization capability of the model, ultimately leading to poor fitting of the objective distribution. Therefore, the nonlinear capacity of the BONN with 12 block size appears to be optimal for the MNIST dataset classification task. Additionally, we also study the influence of the different population size $h$ on the performance of the CMA-ES based on BONN. It can be found from Fig. 2(b) that the avg10acc of the BONN increases correspondingly (92.139% for $h$ =40, 92.455% for $h$ =60, 92.956% for $h$ =80 and 93.185% for $h$ =100) as the population size increases. This trend can be attributed to the fact that the large population size enhances the global searching ability of the CMA-ES based on BONN, leading to superior performance. Thus, for MNIST dataset, block size of BONN and population size $h$ are set as 12 and 100, respectively, in our following CAP algorithm.

The prune strategy applied in our proposed CAP algorithm can significantly decrease the trainable parameters in our BONN but also bring degradation to the performance of our BONN. Thus, it is necessary to verify the effectiveness of the CMA-ES employing on the training of BONN before using our proposed CAP algorithm. Here, to illustrate the effectiveness of the CMA-ES, we employ our previously reported gradient-based BAT algorithm [14] to train the BONN with an identical architecture. As shown in Fig. 2(c), the avg10acc of the BAT algorithm can reach 93.458% (green solid line), which is close to the avg10acc of the CMA-ES (purple solid line, 93.185%). Specially, the confusion matrix of the 100th generation of the CMA-ES is shown in Fig. 2(d). This suggests that the performance of the CMA-ES is similar to that of the BAT algorithm. Moreover, to demonstrate the rationality of our BONN, which is without external phase shifters and consists of only BUMLs, we firstly compare its performance with a BONN which is composed of BUMLs with the external phase shifters. This BONN is set with the same architecture with our BONN (784-10) and is also trained separately by the CMA-ES and BAT algorithm for the MNIST dataset classification task. The calculated results are shown in Fig. 2(c). It can be found that the performance of the CMA-ES (93.444%) is also similar to that of the BAT algorithm (94.116%), which further proves the effectiveness of the CMA-ES. Besides, the avg10acc of our BONN without the external phase shifters is slightly lower than that of the BONN with the external phase shifters, which can primarily be attributed to the fact that the parameters of our BONN are almost half of those of the BONN with the external phase shifters. However, performance degradation of our BONN isn't significant. Further, we construct a block artificial neural network (BANN) with an identical architecture to the previously described BONN (784-10) while using a traditional Adam optimizer [30]. It can be clearly seen from Fig. 2(c) that the avg10acc of our BONN based on CMA-ES (93.185%) is slightly lower than avg10acc of the BANN (94.654%). This is because that our BONN can't express the arbitrary real-valued weight matrix. Our BONN only uses almost 25% of parameter numbers of the BONN with the external phase shifters, which can implement the arbitrary real-valued weight matrix, but can achieve similar performance as the BANN. In other words, substituting the BONN with the external phase shifters, which can implement the arbitrary real-valued weight matrix, with our BONN is an efficient way to reduce both the system area and the required trainable parameters.

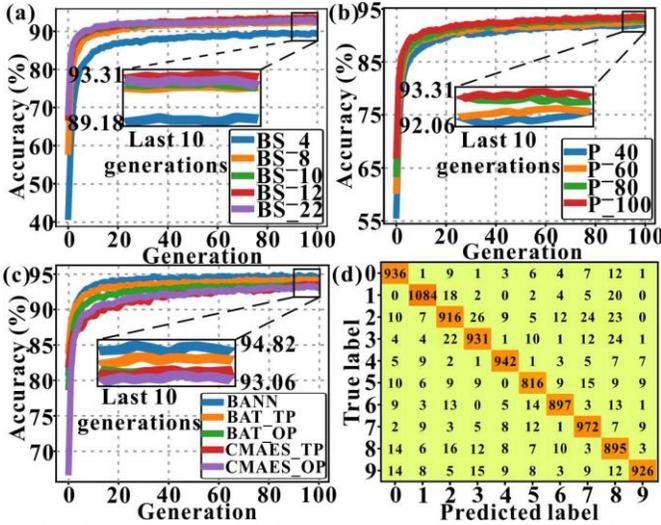

Fig. 2. (a) The accuracy of the BONN with 4 block size (BS_4, blue solid line), 8 block size (BS_8, orange solid line), 10 block size (BS_10, green solid line), 12 block size (BS_12, red solid line) and 22 block size (BS_22, purple solid line) based on CMA-ES for the MNIST dataset classification task. Here, the population size is set as 100. (b) The accuracy of the BONN with 40 population size(P_40, blue solid line), 60 population size (P_60, orange solid line), 80 population size (P_80, green solid line) and 100 population size (P_100, red solid line) based on the CMA-ES. (c) The accuracy of BANN (blue solid line), BONN with the external phase shifters based on BAT algorithm (BAT_TP, orange solid line), our BONN without the external phase shifters based on BAT algorithm (BAT_OP, green solid line), BONN with the external phase shifters based on CMA-ES (CMAES_TP, red solid line) and our BONN without the external phase shifters based on CMA-ES (CMAES_OP, purple solid line). (d) The confusion matrix of the 100th generation of CMAES_OP shows an accuracy of 93.15%.

To further illustrate the effectiveness of the CMA-ES, we apply it to the more complex Fashion-MNIST dataset classification task [31], where each image represents a clothing item rather than a simple number. Like the MNIST dataset, Fashion-MNIST dataset also comprises 60000 training grayscale images with dimensions of 28 × 28 pixels (10 categories) and 10000 testing images. To classify these images, we construct a BONN with an architecture of 10C5-10FC and using the proposed CMA-ES to train this network. Here, 10C5 denotes a convolution layer with 5 ×5 kernel and 10 feature maps, and 10FC represents a fully connected layer with 10 neurons. Moreover, this BONN doesn't include the external phase shifters. To implement convolutional layers in the BONN, we employ the widely-used tensor unrolling method (im2col) [32] to convert convolution operations into matrix multiplication. Firstly, we study the impact of the different block size and population size $h$ on the performance of the CMA-ES based on BONN. It can be found from Fig. 3(a) that the avg10acc of the BONN with 12 block size can reach 85.231%, which is higher than that of the BONN with other block sizes. Moreover, as shown in Fig. 3(b), we can also find that the avg10acc of the BONN increases correspondingly as the population size increases. Thus, for Fashion-MNIST dataset, block size of BONN and population size $h$ are set as 12 and 120, respectively, in our following CAP algorithm. Then we use BAT algorithm to train the BONN with an identical architecture (10C5-10FC). As shown in Fig. 3(c), the avg10acc of the BAT algorithm can reach 85.75%, which is close to the avg10acc of the CMA-ES (85.231%). Specially, the confusion matrix of the 100th generation of the CMA-ES is shown in Fig. 3(d). Further, we construct a BANN with an identical architecture to the previously described BONN while using a traditional Adam optimizer. It can be clearly seen from Fig. 3(c) that the avg10acc of our BONN based on CMA-ES (85.231%) is also close to avg10acc of the BANN (87.289%). This further illustrates the effectiveness of the CMA-ES and suggests that substituting the BONN with the external phase shifters, which can implement the arbitrary real-valued weight matrix, with our BONN is an efficient way to reduce both the system area and the required trainable parameters.

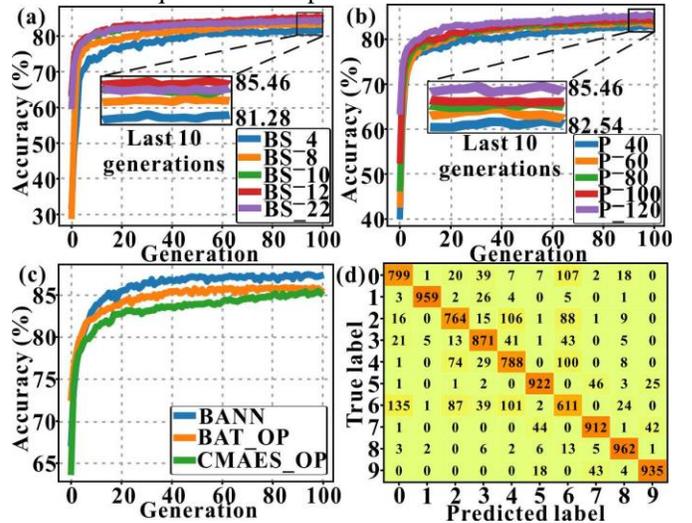

Fig. 3. (a) The accuracy of the BONN with 4 block size (BS_4, blue solid line), 8 block size (BS_8, orange solid line), 10 block size (BS_10, green solid line), 12 block size (BS_12, red solid line) and 22 block size (BS_22, purple solid line) based on CMA-ES for the Fashion-MNIST dataset classification task. Here, the population size is set as 120. (b) The accuracy of the BONN with 40 population size(P_40, blue solid line), 60 population size (P_60, orange solid line), 80 population size (P_80, green solid line), 100 population size (P_100, red solid line) and 120 population size (P_120, purple solid line)based on the CMA-ES. (c) The accuracy of BANN (blue solid line), BONN without the external phase shifters based on BAT algorithm (BAT_OP, orange solid line) and our BONN without the external phase shifters based on CMA-ES (CMAES_OP, green solid line). (d) The confusion matrix of the 100th generation of CMAES_OP shows an accuracy of 85.23%.

## 3.2. Image Classification Based on CAP Algorithm

In the previous part, we have studied the impact of the different block size and population size on the performance of the CMA-ES based on BONN. Because CAP algorithm uses attention mechanism to prune BONN, the remaining hyper-parameter $S$, which represents the ratio of the number of iteration (MaxGen1) required to train the attention coefficient $\alpha$ to the total number of iteration (MaxGen1+MaxGen2), need to be further studied. To explore the impact of the hyper-parameter $S$ on the performance of our proposed CAP algorithm, we build a BONN which has an architecture of 784-10 and is composed of BUMLs without the external phase shifters. This BONN is trained by the CAP algorithm for MNIST dataset classification task. Here, the total number of iteration (MaxGen1+MaxGen2) is set to 100. It can be found from Fig. 4(a) that the avg10acc of the BONN can reach 89.896% (orange solid circle) when $S$=40%, which is higher than that of the BONN when $S$ is set to other values. This characteristic can be attributed to the fact that a too small value of $S$ can lead to relatively inaccurate attention coefficients, which may mistakenly prune weight matrix blocks that have a significant impact on the results. On the other hand, a too large value of $S$ can result in insufficient training of the pruned BONN. Therefore, when $S$ is set to 40%, the BONN based on the CAP algorithm has the best performance for MNIST dataset.

To illustrate the effectiveness of our proposed CAP algorithm based on BONN, we employ the CAP algorithm and RPC algorithm to train the BONN for comparison, respectively. Here, RPC algorithm randomly prunes a portion of the matrix blocks in BONN and then trains the remaining matrix blocks using the CMA-ES. We study the impact of predefined proportion of pruned parameters $R$ on the performance of the CAP algorithm and RPC algorithm based on BONN. Notably, the performance degradation of less than 5% caused by parameter pruning is generally considered acceptable [33]. It can be found from Fig. 4(b) that the avg10acc of both CAP algorithm and RPC algorithm tends to decline as predefined proportion $R$ increases, but the avg10acc of CAP algorithm declines more slowly. Besides, we can also find that CAP algorithm can prune 60% of the parameters with a performance degradation of 3.289%. These suggest that our proposed CAP algorithm is effective for BONN simplification. Further, as the inset histogram shown in Fig. 4(b), our BONN based on CAP algorithm only uses almost 10% of parameter numbers of the BONN with the external phase shifters, which can implement the arbitrary real-valued weight matrix. This indicates that our proposed CAP algorithm is significant for BONN.

In addition, the performance degradation of the BONN is often attributed to parameter imprecisions in the photonic devices. These imprecisions can be classified into two primary types: static error mainly caused by the imperfect 3-dB directional coupler and dynamic error mainly caused by the thermal crosstalk between on-chip phase shifters and its unstable electrical control [34]. The static error $\delta_C$ can be calculated by measuring the extinction ratio $E$ of the MZI in the crossbar state, where $\delta_C = 10^{-E/20}$. Notably, the coupling coefficient error remains fixed after the photonic device is manufactured and is not affected by the external environment. On the other hand, the dynamic error $\delta_P$ can be modeled as a random Gaussian distributed variable $G_P(\mu, \sigma)$ where the expectation $\mu = 0$ and the standard deviation $\sigma = \sigma_P$, and can be added directly to the original phase $\theta$. To illustrate the stronger robustness of our proposed CAP algorithm based on BONN, we firstly use CAP algorithm and

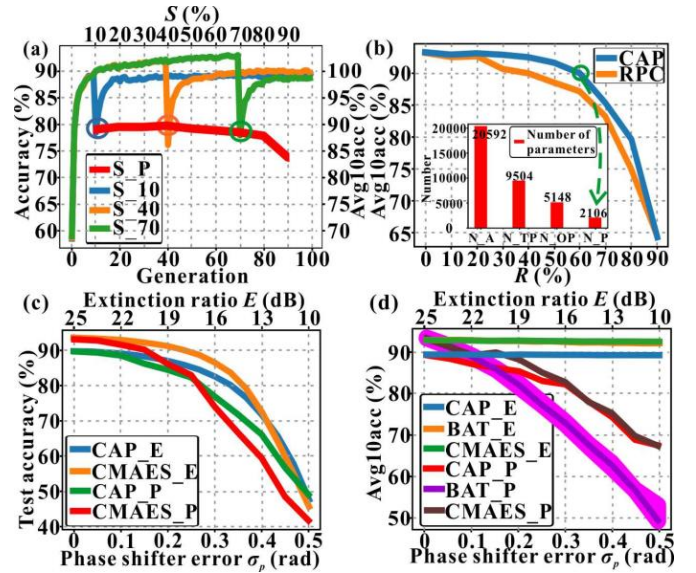

Fig. 4. (a) The avg10acc of the BONN based on the CAP algorithm when $S$ is set to different values (S_P, red solid line), and the accuracy of the BONN based on the CAP algorithm when $S$=10 (S_10, blue solid line), 40 (S_40, orange solid line), 70 (S_70, green solid line) for the MNIST dataset classification task, respectively. (b) The avg10acc of the BONN trained separately by the CAP algorithm (blue solid line) and RPC algorithm (orange solid line) when $R$ is set to different values. Here, the inset histogram shows the numbers of training parameters of the BONN with the external phase shifters, which can implement the arbitrary real-valued weight matrix (N_A), the BONN which is composed of BUMLs with the external phase shifters (N_TP), our BONN which is composed of BUMLs without the external phase shifters (N_OP), and our pruned BONN based on the CAP algorithm (N_P), respectively. (c) The test accuracy of the BONN trained separately by the CAP algorithm (CAP_E, blue solid line) and CMA-ES (CMAES_E, orange solid line) in the case of different extinction ratios of directional coupler, and the test accuracy of the BONN trained separately by the CAP algorithm (CAP_P, green solid line) and CMA-ES (CMAES_P, red solid line) in the case of different standard deviations of phase shifter error. (d) The avg10acc of the BONN trained separately by CAP algorithm (CAP_E, blue solid line), BAT algorithm (BAT_E, orange solid line) and CMA-ES (CMAES_E, green solid line) in the case of different extinction ratios of directional coupler, and the avg10acc of the BONN trained separately by CAP algorithm (CAP_P, red solid line), BAT algorithm (BAT_P, purple solid line) and CMA-ES (CMAES_P, brown solid line) in the case of different standard deviations of phase shifter error. The sand-purple area represents the variation in the performance of the BAT algorithm when training the noisy BONN over five trials.

CMA-ES to train the BONN without noise for 100 generations based on the MNIST dataset classification task, respectively. Then we test the impact of static error and dynamic error on these two trained BONNs. It can be found from Fig. 4 (c) that the test accuracy of both CAP algorithm and CMA-ES tends to decline as static error and dynamic error increases, but the test accuracy of CAP algorithm declines more slowly. Specifically, the test accuracy of the BONN trained by CAP algorithm is 1.9% higher than CMA-ES when the extinction ratio of the directional coupler is 10, and 7.35% higher when the standard deviation of the phase shifter error is 0.5. In addition, we use CAP algorithm, BAT algorithm and CMA-ES to train the BONN with static error and dynamic error for 100 generations, respectively. It can be obviously found from Fig. 4(d) that the avg10acc of CAP algorithm, BAT algorithm and CMA-ES also tends to stability as static error increases. This suggests that static error can be calibrated through online training of algorithms. In other words, the impact caused by static error can be ignored during online training. It can be also found from Fig. 4(d) that the avg10acc of CAP algorithm, BAT algorithm and CMA-ES also tends to decline as the standard deviation of phase shift error increases, but the

avg10acc of CAP algorithm declines the most slowly. Specifically, the avg10acc of CAP algorithm (67.569%) is 18.074% higher than that of BAT algorithm (49.495%) at a standard deviation of 0.5. These suggest our proposed CAP algorithm based on BONN exhibits stronger robustness compared with BAT algorithm and CMA-ES.

Moreover, to further illustrate the effectiveness of our proposed CAP algorithm based on BONN, we build a BONN which has an architecture of 10C5-10FC and is composed of BUMLs without the external phase shifters for Fashion-MNIST dataset classification task. Firstly, we explore the impact of the hyper-parameter $S$ on the performance of our proposed CAP algorithm. Here, the total number of iteration (MaxGen1+ MaxGen2) is set to 100. It can be found from Fig. 5(a) that the BONN can achieve the best performance (80.538%, orange solid circle) when $S$=40%. Then we use the CAP algorithm and RPC algorithm to train the BONN and study the impact of predefined proportion $R$ on the performance of the CAP algorithm and RPC algorithm based on BONN. It can be found from Fig. 5(b) that the avg10acc of CAP algorithm also declines more slowly, which is same as the phenomenon in the MNIST dataset classification task. Besides, we can find that CAP algorithm can prune 80% of the parameters with a performance degradation of 4.693%. These further suggest our proposed CAP algorithm is effective for BONN simplification.

To further illustrate the stronger robustness of our proposed CAP algorithm based on BONN, we also use CAP algorithm and CMA-ES to train the BONN without noise for 100 generations based on the Fashion-MNIST dataset classification task, respectively. Notably, the parameters of the BONN trained by CAP algorithm are pruned by 80%. Then we test the impact of static error and dynamic error on these two trained BONNs. It can be found from Fig. 5(c) that the test accuracy of CAP algorithm also declines more slowly compared with CMA-ES. Specifically, the test accuracy of the BONN trained by CAP is 2.09% higher than CMA-ES when the extinction ratio of the directional coupler is 10, and 8.99% higher when the standard deviation of the phase shifter error is 0.5. Due to the randomness and uncertainty of dynamic error, the BONN with dynamic error is more challenging to be trained compared to the BONN with static error. Therefore, we use CAP algorithm, BAT algorithm and CMA-ES respectively to only train the BONN with dynamic error for 100 generations. It can be obviously found from Fig. 5(d) that the avg10acc of CAP algorithm also declines the most slowly. Notably, the avg10acc of the BONN trained by CAP algorithm (56.519%) is 11.843% higher than that of BAT (44.676%) and 4.159% higher than that of CMA-ES (52.36%) when the standard deviation of phase shifter error is 0.5. These further suggest our proposed CAP algorithm based on BONN exhibits stronger robustness compared with BAT algorithm and CMA-ES and this effect is even more pronounced in more complex Fashion-MNIST task.

### 3.3. Image Classification Based on Experiment

As shown in Fig. 6(a), we set up an experimental platform [35]. The system utilizes a polarization-maintaining (PM) laser (CoBrite-DX4, ID Photonics) with a peak output power of 15.5 dBm. In actual experiment, the output power of the laser is set to 12 dBm. The PIC is manufactured using a 180 nm CMOS process technology, and incorporates a 4×4 MZI array designed based on rectangle decomposition architecture [22]. Temperature control is achieved through a thermo-electric cooler (TEC), which dynamically adjusts the current based on real-time temperature read from a thermistor embedded beneath the chip. The output power of the PIC is measured using a PD array (EBR370005-02, EXALOS). The field programmable gate array (FPGA), controlled by EC, configures the voltage on the PIC's electrical interface and samples the computational results. Throughout the experiment, the thermo-optical modulators are operated at a fixed modulation rate of 1 kHz.

To verify the practicality of our proposed CAP algorithm, we deploy the CAP algorithm on this experimental platform and use the CAP algorithm to train the BONN based on the 4×4 MZI array online for MNIST dataset classification task. Considering the data processing speed of the experimental platform, we randomly select 100 images from MNIST dataset and use convolutional layers with pre-trained weights and flatten operation to reduce the dimensionality of the 28 ×28 images to 1 ×32. These processed images are used as the dataset for the experiment. Firstly, we study the impact of population size $h$, hyper-parameter $S$ and predefined proportion $R$ on the performance of the CAP algorithm based on BONN in simulation. It can be found from Fig. 6(b) that the avg10acc of the BONN (blue solid line) increases correspondingly (97% for $h$=20) as the population size increases, and the BONN can achieve the best performance (92.1%, orange solid line) when $S$=40%. It can be also found from Fig. 6(b) that CAP algorithm can prune 60% of the parameters with a performance degradation of 4.9% (green solid line). Thus, for simplified MNIST dataset, $h$, $S$ and $R$ are set as 20, 40% and 60%, respectively, in our following experiment. Then we use the CAP algorithm to experimentally train the BONN without the external phase shifters and the BONN with the external phase shifters, respectively. Notably, the voltage values corresponding to the external phase shifters in the PIC are set to 0 for the BONN without the external phase shifters. It can be found from Fig. 6(c) that the avg10acc of the BONN without the external phase shifters (green solid line) can reach 88.5% (92.1% in simulation without noise, blue solid line) and the avg10acc of the BONN with the external phase shifters (red solid line) can reach 91.5% (92.3% in simulation without noise, orange solid line). These suggest that the performance of the CAP algorithm in experiment is similar to that in simulation and demonstrate the practicality of our proposed CAP algorithm. Notably, the performance of BONN without the external phase shifters shows a slightly larger difference between simulation and experiment compared with that of BONN with the external phase shifters. This characteristic can be attributed to the fact that the BONN without the external phase shifters may be affected by external phase shifter noise in actual experiment, but its relatively small number of training parameters can't effectively calibrate the noise introduced by external phase shifters.

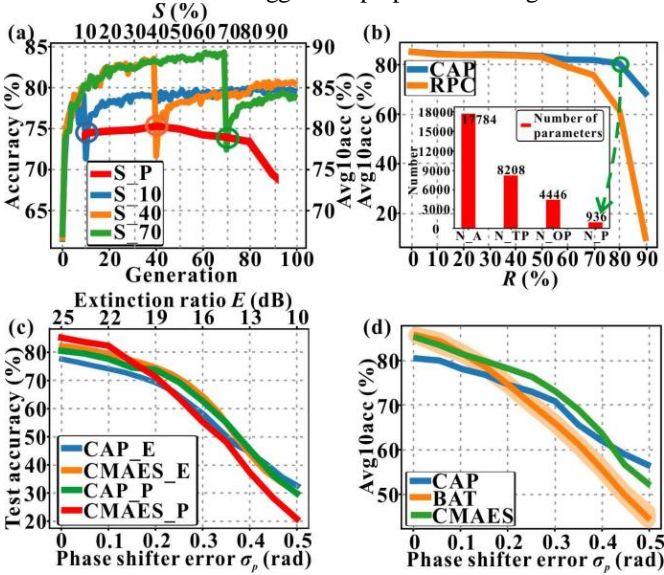

Fig. 5. (a) The avg10acc of the BONN based on the CAP algorithm when $S$ is set to different values (S_P, red solid line), and the accuracy of the BONN based on the CAP algorithm when $S$=10 (S_10, blue solid line), 40 (S_40, orange solid line), 70 (S_70, green solid line) for the Fashion-MNIST dataset classification task, respectively. (b) The avg10acc of the BONN trained separately by the CAP algorithm (blue solid line) and RPC algorithm (orange solid line) when $R$ is set to different values. Here, the inset histogram shows the numbers of training parameters of the BONN with the external phase shifters, which can implement the arbitrary real-valued weight matrix (N_A), the BONN which is composed of BUMLs with the external phase shifters (N_TP), our BONN which is composed of BUMLs without the external phase shifters (N_OP), and our pruned BONN based on the CAP algorithm (N_P), respectively. (c) The test accuracy of the BONN trained separately by the CAP algorithm (CAP_E, blue solid line) and CMA-ES (CMAES_E, orange solid line) in the case of different extinction ratios of directional coupler, and the test accuracy of the BONN trained separately by the CAP algorithm (CAP_P, green solid line) and CMA-ES (CMAES_P, red solid line) in the case of different standard deviations of phase shifter error. (d) The avg10acc of the BONN trained separately by CAP algorithm (CAP, blue solid line), BAT algorithm (BAT, orange solid line) and CMA-ES (CMAES, green solid line) in the case of different standard deviations of phase shifter error. The sand-orange area represents the variation in the performance of the BAT algorithm when training the noisy BONN over five trials.

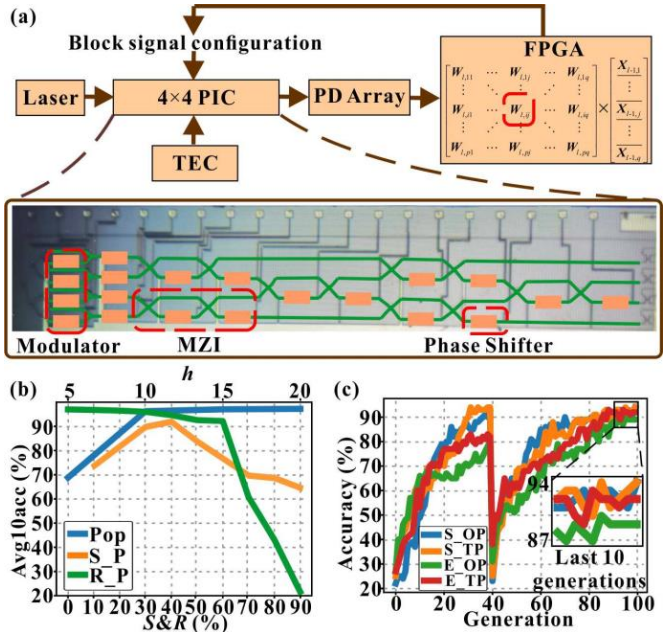

Fig. 6. (a) The experimental platform based on 4×4 MZI array. (b) The avg10acc of the BONN when $h$ is set to different values (Pop, blue solid line), the avg10acc of the BONN based on the CAP algorithm when $S$ is set to different values (S_P, orange solid line), and the avg10acc of the BONN based on the CAP algorithm when $R$ is set to different values (R_P, green solid line), respectively. (c) The accuracy of the BONN without the external phase shifters in simulation (S_OP, blue solid line), the accuracy of the BONN with the external phase shifters in simulation (S_TP, orange solid line), the accuracy of the BONN without the external phase shifters in experiment (E_OP, green solid line), and the accuracy of the BONN with the external phase shifters in experiment (E_TP, red solid line), respectively.

## 4. Conclusion

In conclusion, by pruning blocks that have minimal impact on the results and directly optimizing the individuals in the population, we propose a novel on-chip CAP algorithm for large-scale BONNs. After demonstrating that the CMA-ES based on the BONNs with 12 block size and the largest possible population can achieve higher performance, the BONNs trained by our algorithm are applied in image classification tasks for MNIST and Fashion-MNIST datasets to demonstrate the effectiveness of our proposed algorithm. The calculated results demonstrate that the CAP algorithm can prune 60% and 80% of the parameters while only degrades the performance of 3.289% and 4.693% for MNIST and Fashion-MNIST datasets, respectively. Considering the influence of dynamic noise in phase shifters, our proposed CAP algorithm (performance degradation of 22.327% for MNIST dataset and 24.019% for Fashion-MNIST dataset utilizing a poor fabricated chip and electrical control with a standard deviation of 0.5) exhibits strongest robustness compared with both our previously reported BAT algorithm (43.963% and 41.074%) and the CMA-ES (25.757% and 32.871%), respectively. The calculated results also demonstrate that we can substitute the traditional BONN, which can implement arbitrary real-valued matrices, with our BONN composed of BUML without external phase shifters to reduce both the system area and the required trainable parameters by approximately 75% while almost maintaining the performance. Moreover, when 60% of the parameters are pruned, the CAP algorithm based on the BONN without the external phase shifters and that based on the BONN with the external phase shifters can realize 88.5% accuracy and 91.5% accuracy in experiment for the simplified MNIST dataset, respectively, which is similar to the simulation result without noise (92.1% and 92.3%). Notably, our proposed CAP algorithm show excellent potential for larger-scale network models and more complex tasks.

## CRediT authorship contribution statement

**Zhiwei Yang:** Conceptualization, Methodology, Software, Writing – original draft, Writing – review & editing. **Zeyang Fan:** Software. **Yihang Lai:** Software. **Qi Chen:** Software. **Tian Zhang:** Conceptualization, Funding acquisition, Resources, Supervision, Writing – review & editing. **Jian Dai:** Project administration. **Kun Xu:** Project administration, Funding acquisition.

## Declaration of competing interest

The authors declare that they have no known competing financial interests or personal relationships that could have appeared to influence the work reported in this paper.

## Acknowledgments

This work was supported by the National Natural Science Foundation of China (62171055, 62135009, 62471062); Fundamental Research Funds for the Central Universities (ZDYY202102); BUPT Innovation and Entrepreneurship Support Program (2021-YC-A110), P. R. China; Super Computing Platform of Beijing University of Posts and Telecommunications.

## Data availability

Data will be made available on request.